\documentclass[AMA,Times1COL]{WileyNJDv5} 
\usepackage{subfigure}
\usepackage{color,array}
\usepackage{tabularx}
\usepackage[section]{placeins}
\usepackage{afterpage}

\usepackage{graphicx}
\usepackage{times}
\usepackage{multirow}
\usepackage{blindtext}
\usepackage{algpseudocode}
\usepackage{bm}
\usepackage{latexsym}
\usepackage{booktabs}
\usepackage{threeparttable}
\usepackage{epsf}
\usepackage{color,colortbl}
\usepackage{babel}
\usepackage{caption}
\renewcommand{\algorithmicensure}{\textbf{Output:}}
\renewcommand{\algorithmicrequire}{\textbf{Input:}}
\usepackage{hyperref}
\hypersetup{
	colorlinks=true,
	linkcolor=blue,
	anchorcolor=blue,
	citecolor=blue}

\usepackage{multirow}
\usepackage{ulem}
\useunder{\uline}{\ul}{}
\usepackage[T1]{fontenc}
\usepackage{amsmath}
\usepackage{color}
\usepackage[misc]{ifsym}

\usepackage{hyphenat}
\articletype{ORIGINAL ARTICLE}%

\renewcommand{\algorithmicrequire}{ \textbf{Input:}} 
\renewcommand{\algorithmicensure}{ \textbf{Output:}} 

\received{Date Month Year}
\revised{Date Month Year}
\accepted{Date Month Year}
\journal{Journal}
\volume{00}
\copyyear{2024}
\startpage{1}

\begin{document}

\title{Peak-Controlled Logits Poisoning Attack in Federated Distillation}

\author[1,2]{Yuhan Tang}
\author[1,2]{Aoxu Zhang}
\author[3,4]{Zhiyuan Wu}
\author[1,2]{Bo Gao}
\author[3]{Tian Wen}
\author[3]{Yuwei Wang}
\author[3]{Sheng Sun}

\authormark{Y. Tang \textsc{et al.}}
\titlemark{Peak-Controlled Logits Poisoning Attack in Federated Distillation}

\address[1]{Engineering Research Center of Network Management Technology for High-Speed Railway of Ministry of Education, School of Computer Science and Technology, Beijing Jiaotong University, Beijing, China}
\address[2]{Collaborative Innovation Center of Railway Traffic Safety, Beijing Jiaotong University, Beijing, China}
\address[3]{Institute of Computing Technology, Chinese Academy of Sciences, Beijing, China}
\address[4]{University of Chinese Academy of Sciences, Beijing, China}


\corres{Bo Gao \email{bogao@bjtu.edu.cn}}



\abstract[Abstract]{Federated Distillation (FD) offers an innovative approach to distributed machine learning, leveraging knowledge distillation for efficient and flexible cross-device knowledge transfer without necessitating the upload of extensive model parameters to a central server. While FD has gained popularity, its vulnerability to poisoning attacks remains underexplored. To address this gap, we previously introduced FDLA (Federated Distillation Logits Attack), a method that manipulates logits communication to mislead and degrade the performance of client models. However, the impact of FDLA on participants with different identities and the effects of malicious modifications at various stages of knowledge transfer remain unexplored. To this end, we present PCFDLA (Peak-Controlled Federated Distillation Logits Attack), an advanced and more stealthy logits poisoning attack method for FD. PCFDLA enhances the effectiveness of FDLA by carefully controlling the peak values of logits to create highly misleading yet inconspicuous modifications. Furthermore, we introduce a novel metric for better evaluating attack efficacy, demonstrating that PCFDLA maintains stealth while being significantly more disruptive to victim models compared to its predecessors. Experimental results across various datasets confirm the superior impact of PCFDLA on model accuracy, solidifying its potential threat in federated distillation systems.}


\jnlcitation{\cname{%
\author{Y. Tang},
\author{Z. Aoxu},
\author{W. Zhiyuan},
\author{G. Bo},
\author{W. Tian}, and
\author{W. Yuwei}
\author{S. Sheng}}.
\ctitle{On simplifying ‘incremental remap’-based transport schemes.} \cjournal{\it J Comput Phys.} \cvol{2024;00(00):1--18}.}

\maketitle
\footnotetext{\textbf{Abbreviations:} FL,Federated Learning; FD,Federated Distillation; KD,Knowledge Distillation; FDLA,Federated Distillation Logits Attack; PCFDLA,Peak-Controlled Federated Distillation Logits Attack.}

\renewcommand\thefootnote{\fnsymbol{footnote}}.
\setcounter{footnote}{1}

\section{Introduction}
Federated Learning (FL) \citep{yang2019federated,mcmahan2017communication,li2020federated} has emerged as a prominent distributed machine learning paradigm that fully utilizes local data and computation power of mobile clients for training deep learning models in a privacy-preserving manner. Despite its significant advantages over traditional machine learning methods, FL encounters several challenges, including low communication efficiency \cite{kairouz2021advances,wu2022communication,wang2022communication} and device heterogeneity \cite{tan2022towards,tak2020federated}. Federated Distillation (FD) \cite{jeong2018communication,itahara2021distill,li2024federated,wu2023fedcache,li2019fedmd,wu2023fedict}, a variant of FL, addresses these issues by exchanging only model outputs (knowledge) during training, avoiding the cumbersome transmission of model parameters. FD effectively combines FL and Knowledge Distillation (KD) \cite{hinton2015distilling,wang2021knowledge,wu2021spirit} to meet the demands of efficient communication and heterogeneous device environments, showcasing considerable advantages and applications in various emerging fields.

Despite the remarkable success of FD, previous research efforts have focused on performance improvement \cite{wu2023fedict,deng2023hierarchical,zhang2021parameterized}, communication optimization \cite{itahara2021distill,wu2023fedcache,jeong2018communication}, or system compatibility \cite{he2020group,cho2022heterogeneous,li2019fedmd,cheng2021fedgems,wu2023agglomerative,huang2022learn}, while overlooking security challenges. Specifically, FD is vulnerable to logits poisoning attacks, where the knowledge uploaded can be easily modified without adequate protection \cite{tang2024logits}. This vulnerability arises because the central server cannot directly monitor the local knowledge generation during FD training, allowing malicious participants to interfere with the model training process. Although pollution attacks in FL have been extensively studied \cite{sun2021data, yang2023clean, gong2022backdoor, cao2022mpaf, 10287523}, the field of poisoning attacks against FD systems remains underexplored. Furthermore, FD's unique characteristics distinguish it from traditional FL, making previous FL attack methods unsuitable for FD. Consequently, there is a pressing need for targeted research on attacks specific to FD.

To fill this gap, we propose FDLA (Federated Distillation Logits Attack) \cite{tang2024logits} in our previous work, which is a logits manipulation attack method specifically designed for Federated Distillation (FD). FDLA strategically modifies the transmitted logits sent to the server to mislead the optimization process of the local client models. The proposed manipulation goal is to generate false logits with higher confidence, aligning the model output with incorrect classification results. This method not only causes the model to make erroneous predictions but also injects a high degree of confidence into these incorrect predictions. FDLA significantly alters the transmission of logits in the FD process, subtly and effectively intervening in model training. Furthermore, the FDLA attack method is executed without the need to modify any private data, paving a new path for interfering with the model training process. However, FDLA has some notable limitations: 1. Due to the direct FDLA attack at the model output stage, the model itself receives incorrect knowledge, causing it to lose the ability to identify correct results after multiple training rounds. Ultimately, without the attacker having the capability to predict correct answers in advance, the model can only modify the output each time without independently determining the correct answer for each test. 2. FDLA has not explored the precision of the attacker's and honest participants' models, so the performance analysis of these two types of models before and after the attack remains unknown.

In this paper, we propose a new logits attack method to address the shortcomings of FDLA, called Peak-Controlled Federated Distillation Logits Attack (PCFDLA). PCFDLA adjusts the confidence of the logits closest to the correct answer target to a preset higher value, while setting the confidence of other non-target types to lower values. For instance, when interfering with the training process of a dataset containing images of cats, the attacker adjusts the confidence of the label "deer" to make it the most credible output result, thereby misleading other models into thinking that the cat images are of deer. PCFDLA only modifies the uploaded knowledge incorrectly, ensuring that the knowledge used for its local training remains correct. It inherits all the advantages of FDLA and addresses the shortcoming of FDLA's inability to retain the capability to judge correct results. By modifying the knowledge during the upload phase, it gains the ability to predict the correct answer for each sample. To better evaluate the performance of PCFDLA in attacking the precision of the models of both the attacker and honest participants, we also propose a new evaluation metric based on FDLA's metrics. This new metric can test the impact of the attack on the accuracy of both the attacker's and non-attacker's models and assess the extent of damage caused by all attack methods. This allows for a more detailed understanding of the effectiveness of the attack method.

In general, our contributions can be summarized as follows:
\begin{itemize}
    \item 
    We propose a novel method, PCFDLA, for logits poisoning attack in Federated Distillation (FD). PCFDLA manipulates the confidence levels of logits during training, enableing the model to more accurately select incorrect but seemingly reasonable results to achieve the purpose of deception.
    \item 
    We propose a new metric to evaluate the effectiveness of different attack methods by recording the model accuracy of both the victims and malicious attackers before and after the attack. Our proposed metric can more clearly and comprehensively show the extent of damage caused to the victims by different attacks.
    \item 
    We evaluated our proposed method with multiple settings on the CINIC-10, CIFAR-10, and SVHN datasets. The results show that PCFDLA achieves a more significant reduction in model accuracy compared to FDLA and baseline algorithms. Additionally, we proposed more detailed and extensive experiments than FDLA, including testing the recognition accuracy of both the attackers and honest participants before and after poisoning, as well as testing the attack intensity of PCFDLA.
\end{itemize}
\section{Related Work}
\subsection{Poisoning Attack in Federated Learning}
    In recent years, researchers have extensively focused on poisoning attack strategies that compromise the performance of federated learning models, proposing various countermeasures.
These strategies include implicit gradient manipulation to attack sample data \cite{sun2021data}, unlabeled attacks \cite{yang2023clean}, backdoor attacks by replacing components of the model \cite{gong2022backdoor}, and the introduction of fake clients to distort the global model's learning process \cite{cao2022mpaf}. Additionally, a technique called VagueGAN employs GAN models to generate blurry data for training toxic local models, which are then uploaded to servers to degrade the performance of other local models \cite{10287523}. These attack methods have been well-studied and have influenced traditional defense mechanisms. While these methods show excellent performance in terms of attack efficiency and covert behavior within classical federated learning frameworks, they are not directly applicable to the FD scenario, leaving room for improvement and further exploration. Therefore, it is necessary to further explore and develop attack and defense strategies tailored to the characteristics of the FD scenario. Previous research on FDLA \cite{tang2024logits} proposed tampering with client-side outputs of knowledge results to confuse the aggregation of knowledge on servers. This approach aimed to contaminate the knowledge received by other clients and potentially degrade overall model performance. However, this method fails to effectively mislead models into providing highly accurate but incorrect answers, occasionally presenting alternative incorrect answers. This paper addresses the shortcomings of existing work, integrating considerations of both FL and FD scenario characteristics, identifying potential vulnerabilities introduced by tampering with knowledge, and filling critical gaps in the literature. The paper introduces PCFDLA, an attack method with precise misleading capabilities and controllability, addressing the deficiencies identified in previous research and further exploring vulnerabilities in federated distillation.

\subsection{Federated Distillation}
As an emerging paradigm in federated learning, Federated Distillation (FD) applies knowledge distillation for model optimization and has garnered increasing attention in recent years \cite{li2024federated}. Specifically, FD effectively overcomes system heterogeneity through collaborative optimization and knowledge sharing at the client level \cite{li2019fedmd,wu2022exploring,pan2024fedcache}. By guiding local models towards different learning objectives, FD meets the personalized needs of clients \cite{zhang2021parameterized,wu2023fedict}. It reduces the communication bandwidth required for training by transmitting only lightweight model outputs \cite{wu2023fedcache,sattler2021cfd,itahara2020distillation}. Through model-agnostic interaction protocols, it enables the training of larger models on the server \cite{wu2023agglomerative,cheng2021fedgems,cho2022heterogeneous}. Furthermore, by conducting self-distillation on the client side, FD mitigates client drift caused by data heterogeneity or incremental characteristics \cite{yao2023f,wu2024federated}. However, existing methods primarily focus on performance improvements of FD algorithms and lack exploration of potential security threats in FD \cite{mora2022knowledge,wu2023survey}. Although FDLA \cite{tang2024logits} investigates the impact of logits poisoning attacks in FD, participants with different identities and the effects of malicious modifications at various stages of knowledge transfer remain unexplored.
\section{Approach}
\begin{table}
    \centering
    \caption{Main notations with descriptions.}
    \label{notation}
    \begin{tabular}{c|c}
        \hline
        Notation & Description \\
        \hline
        $k$ & Number of clients \\
        $kv$ & total number of attacked client devices \\
        $D_k$ & Local data \\
        $L_{CE}(\cdot)$ & Cross-entropy loss function \\
        $ZS$ & Global knowledge \\
        $L_{KD}$ & Distillation loss function \\
        \( \beta \) & Knowledge distillation weight factor \\
        $X_{k}$ & Samples of the $k$-th client \\
        $y_{k}$ & Labels of the $k$-th client samples \\
        $W_{k}$ & Model weights of the $k$-th client \\
        $f(W_k;\cdot)$ & Prediction function of the $k$-th client \\
        \( Z_{k} \) & Extracted knowledge of the $k$-th client \\
        $lr$ & Learning rate \\
        $c_i'$ & $i$-th element of the transformed confidence vector \\
        $c'$ & Transformed confidence vector \\
        $i$ & Index of the confidence vector \\
        $S$ & Perturbation factor \\
        \( r(\cdot) \) & Confidence ranking function \\
        \( t(c_{i}) \) & Transformation function \\
        \( x(\cdot) \) & Confidence tampering function \\
        \( TolAvgAcc \) & average accuracy of all clients \\
        \( VctmAvgAcc \) & the average accuracy of all victims \\
        \hline
    \end{tabular}
\end{table}

\subsection{General Process of Federated Distillation}
In the context of FD, each client $k \in \{1,2,...,K\}$ optimizes its model parameters by combining the cross-entropy loss function $L_{CE}(\cdot)$ on local data $D_k$ and the distillation loss function $L_{KD}$ on the global knowledge $ZS$ provided by the server. The server coordinates the collaborative learning process by aggregating the logits (softmax outputs) from all clients. This iterative training process involves many communication rounds, each consisting of a local optimization phase and a global aggregation phase, ultimately updating the local model parameters. For more details on the main notations used in this paper, see Table \ref{notation}.

Regarding the local optimization phase, each client $k$ optimizes the following objectives:\\
\begin{equation}
\label{J_{k}}
    \mathop {\min }\limits_{{{\rm{W}}_k}} \mathop {J_k}\limits_{(X_k^i,y_k^i)\sim{D_k}} [{J_{CE}} + \beta  \cdot {J_{KD}}],
\end{equation}	
where \( \beta \) is the distillation weighting factor, and \( (X_{k}, y_{k}) \) corresponds to the samples and their associated labels of the local dataset \( D_{k} \). The optimization components \( J_{KD}\) and \( J_{CE}\) can be expressed using the following equations:
\
\begin{equation}
    J_{CE} =  L_{CE}(f(W_{k};X_{k}),y_{k}),
\end{equation}
\begin{equation}
    J_{KD} =  L_{KD}(f(W_{k};X_{k}),ZS),
\end{equation}
where \( W_k \) denotes the model parameters of client $k$, and $f(W_k;\cdot)$ is the prediction function on client $k$.
The server performs the aggregation operation over clients' extracted knowledge \(Z_{k}\), and can be typically formulated as knowledge averaging:
	\begin{equation}\label{aggreation_logits}
	A(Z_{k})=\frac{1}{K}\sum_{k \in \{1,2,...,K\}}Z_{k}
\end{equation}

Upon acquiring global knowledge on the server, it is disseminated to all clients. In response, each client proceeds to update their local models with learning rate $lr$ to align with this shared knowledge, with the process formulated as follows:

\begin{equation}\label{update}
	W_{k} = W_{k} - lr \cdot \nabla_{W_{k}} J_{k}(W_{k}).
\end{equation}


\subsection{FDLA}

\begin{figure}[h]
	\centering
	\includegraphics[scale = 0.5]{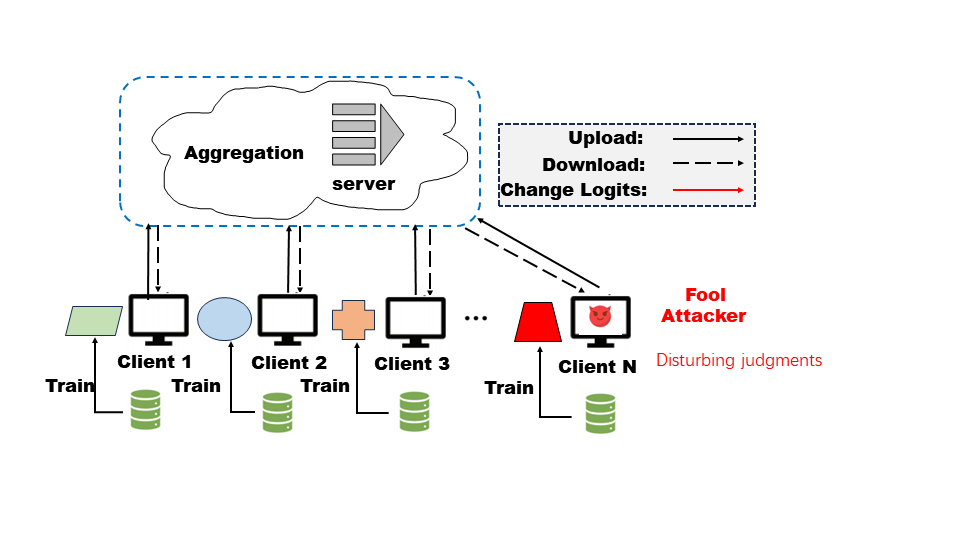}
	\caption{Illustration of how FDLA attackers manipulates uploaded knowledge.}
    \label{FDLA_Logits_Misdirection_Attack_Diagram}
\end{figure}

FDLA is a malicious method specifically designed for Federated Distillation (FD), where malicious clients interfere with the joint learning process of the model by altering the confidence levels of the logits output on local data. "Confidence" refers to the model's certainty in its predictions, indicating the reliability of the model's predictions for a given input. When attackers modify their own model's confidence values, it can lead to decision biases, thereby affecting the final results. Since the training process of federated distillation heavily relies on the output knowledge of the models, with confidence being an important component of this knowledge, modifying the distribution of confidence levels can lead to changes in the model's final output. Intentionally manipulating the rearrangement of confidence values can result in performance changes for both the attacker's model and other models.

Specifically, the client convinces itself to accept the incorrect answer first and uses this incorrect answer for learning, uploading their erroneous knowledge \(Z_{k}\) to the server. This process involves sorting and transforming the logits values \(C\) for each sample, where \(C = [c_{1}, c_{2}, ..., c_{n}]\) contains confidence levels corresponding to each label, with \(c_{i}\) representing the confidence of the \(i\)-th element, assuming \(n\) is the total number of corresponding labels.

The transformation process of FDLA is illustrated in Figure \ref{FDLA_Logits_Misdirection_Attack_Diagram}: First, the confidence values \(c\) are sorted to obtain the sorted indices \(I\). Then, a transformation mapping \(t\) is defined, mapping the element with the highest confidence to the position of the lowest confidence, while the remaining elements are shifted up one position.When the manipulated knowledge \(Z_{k}\) from the malicious client is uploaded to the aggregation server, the server performs the aggregation operation \(A\). The aggregated global knowledge is then distributed to each client, which updates their local models based on this information. However, because the attacker first accepts the incorrect answer, their own model will prioritize this incorrect knowledge, leading to a degradation of their own recognition ability. Although the logits stored on the server are polluted after aggregation and ultimately affect the logits sent to each client, resulting in errors throughout the training process, this is not a smart or efficient approach.

To be more specific,  FDLA execution process is formulated into three steps:
\begin{itemize}
	\item
    [{(\it 1)}]{\it Sort the confidences \(c\) to obtain the sorted indices \(I\).}
	\item
    [{(\it 2)}]{\it Define a transformation mapping \(t\), where \(t[I[1]] = I[n]\), and for \(2 \leq k \leq n\), \(t[I[k]] = I[k - 1]\), where \(I[k]\) is the index of the \(k\)-th highest confidence.}
	\item
    [{(\it 3)}]{\it Apply the transformation mapping \(t\) to the original confidence vector \(c\) to obtain the transformed vector \(c'\).}
\end{itemize}

We express the operation \( CL(c) \) as follows:
\begin{equation}\label{change logits FDLA}
	c^{\prime}_{i}=\left\{
	\begin{array}{ll}
		c_{t[i]} &,  \text{if } r(c_{i}) > 1 \\
		c_{t[i]} &,  \text{if } r(c_{i}) = 1
	\end{array} \right.,
\end{equation}
where \( c^{\prime}_{i} \) is the \( i \)-th element in the transformed confidence vector \( c^{\prime} \), \( r(c_{i}) \) is the rank of \( c_{i} \) in \( c \), and \( t[i] \) is the transformed index.

\subsection{PCFDLA}

\begin{figure}[h]
	\centering
	\includegraphics[scale = 0.5]{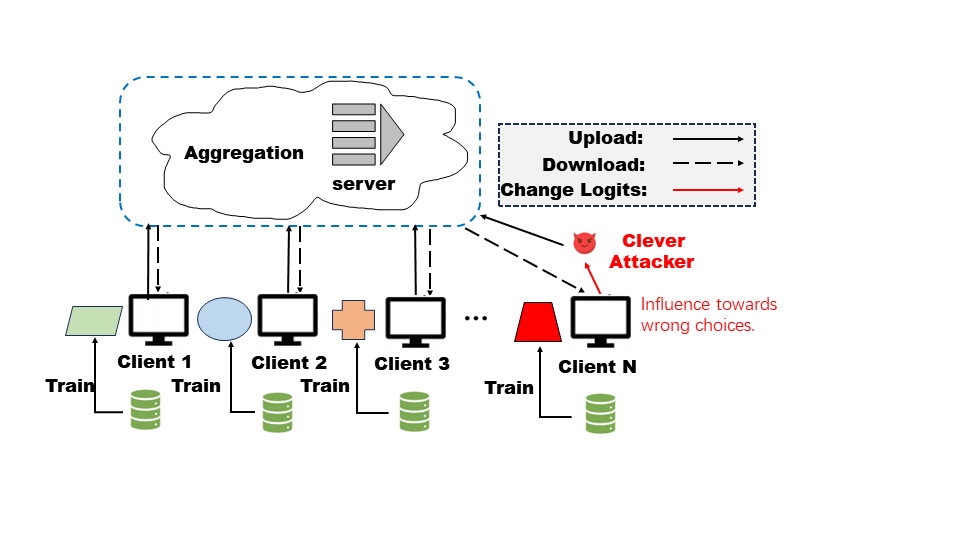}
	\caption{Illustration of how PCFDLA attackers manipulates uploaded knowledge.}
    \label{PCFDLA_Logits_Misdirection_Attack_Diagram}
\end{figure}

Despite the success of FDLA, it did not consider the importance of maintaining the attacker's own ability to accurately judge samples. PCFDLA is an enhanced and controllable malicious attack method specifically designed for FD, and it is smarter than FDLA. Although malicious clients also disrupt the joint learning process of the model by transforming the confidence levels of logits on local data, attackers only modify the uploaded data and do not modify the local training results. More importantly, when modifying confidence values, the attackers are aware of which incorrect answers are most misleading and make the most impactful changes to affect the final result. When this modified answer is used as a reference answer by other models, it may cause the decline of their performance. Specifically, clients do not share their correct answers with others but upload their erroneous knowledge \(Z_{k}\) to the server. This manipulation involves transforming the logits values \(C\) for each sample, where \(C = [c_{1}, c_{2}, ..., c_{n}]\) contains confidence levels corresponding to each label, with \(c_{i}\) representing the confidence of the \(i\)-th element, and \(n\) representing the total number of corresponding labels.

The transformation process of PCFDLA is illustrated in Figure \ref{PCFDLA_Logits_Misdirection_Attack_Diagram}: First, the confidence values \(c\) are sorted to obtain the sorted indices \(I\). Then, a transformation mapping \(t\)is defined, mapping the element with the highest confidence to a set value \(S\), while the remaining elements are mapped to the opposite of the set value. When the manipulated knowledge \(Z_{k}\) from the malicious client is uploaded to the aggregation server, the server performs the aggregation operation \(A\). The aggregated global knowledge is then distributed to each client, which updates their local models based on this information. Since the attacker's model does not first accept and learn this incorrect answer but instead tells everyone the incorrect answer first to gain their acceptance, the attacker will know which incorrect answer is the most misleading each time they encounter a new test and upload the answer. Once all the tests are passed, meaning the logits stored on the server after aggregation are already polluted, and the attacker and other models have accepted the incorrect answer, the attacker then convinces themselves to accept the incorrect answer like the other models, completing the brainwashing of all models and ending the attack, ultimately causing erroneous guidance throughout the training process.

When uploading local data \(D_{k}\) at the \(t\)-th time, the malicious client \(k\) modifies the confidence of the knowledge generated for the \(i\)-th sample \(X_{k}^{i}\) in the current communication round, denoted as \(Z_{k}\). We define a confidence ranking function \( r(·) \) and a confidence tampering function \( x(·) \). The ranking function \( r(c_{i}) \) maps \( c_{i} \) to its rank in the confidence vector (the highest confidence is ranked 1, the second highest is 2, and so on). Then, the transformation function \( x(c_{2}) \) tampers \( c_{2} \) to a new confidence \( S \), where \( S \) is a preset hyperparameter value. Besides mapping the second highest confidence to the preset confidence, other confidences  \( c_{i} \) is a preset hyperparameter value. Besides mapping the second highest confidence to the preset confidence, other confidences \(-S\). This transformation can predict the most similar result as the correct result and completely negate the correctness of other labels to further reduce recognition accuracy.

Our proposed execution process is formulated into three steps:
\begin{itemize}
	\item
    [{(\it 1)}]{\it Sort the confidences \(c\) to obtain the sorted indices \(I\).}
	\item
    [{(\it 2)}]{\it Define a transformation function \(x\), where \( x(c[I[2]],S)\), and for \(1 \leq k \leq n\) except \(k = 2\), \( x(c[I[k],-S)\), where \(I[k]\) is the index of the \(k\)-th highest confidence.}
	\item
    [{(\it 3)}]{\it Apply the transformation function \(x\) to the original confidence vector \(c\) to obtain the transformed vector \(c'\).}
\end{itemize}

We express the operation \( CL(c) \) as follows:
\begin{equation}\label{change logits PAFLDA}
	c^{\prime}_{i}=\left\{
	\begin{array}{ll}
		S &,  \text{if } i = 2 \\
		-S &,  \text{if } i \neq 2
	\end{array} \right.,
\end{equation}
Among them, $c_i'$ is the $i$-th element in the transformed confidence vector $c'$, $i$ is the index of the confidence vector, and $S$ is the preset confidence value.

Originally, the value $c_2$, which was identified as merely the closest to the correct result, is considered the most correct result, while other possibilities $c_i$, including the correct answer $c_1$, are firmly regarded as incorrect. When the malicious attacker's $Z_k$ is uploaded to the aggregation server, the server performs the aggregation operation $A$ as shown in Algorithm \ref{aggregation on server}.

After obtaining the aggregated global knowledge, it is distributed to each client. Subsequently, the clients update their local models based on this newly acquired information, as shown in Algorithm \ref{logits changed on Client k}. During this process, since the logits stored on the server after aggregation have already been polluted, the logits ultimately sent to each client are affected, leading to a misled training process.

\FloatBarrier
\begin{algorithm}[t]
	\renewcommand{\algorithmicrequire}{\textbf{Input:}}
	\renewcommand{\algorithmicensure}{\textbf{Output:}}
	\caption{FDLA or PCFDLA on Client k}
	\label{logits changed on Client k}
	\begin{algorithmic}[1] 
		\Repeat
			\For{client k in K} 
				    \State $W_{k} \leftarrow W_{k} - lr · \nabla_{wk} · J_{k}(W_{k})$; 	according to Eq.(\ref{update})
					\State $Z_{k}$ $\leftarrow$ $f(W_{k},X_{k})$;
					\If{client k is venomous client}
                            \If{algorithm = FDLA}
    					\State $Z_{k}$ $\leftarrow$ CL($X_{k}$);   according to Eq.(\ref{change logits FDLA})
                            \EndIf
                            \If{algorithm = PCFDLA}
                            \State $Z_{k}$ $\leftarrow$ CL($c$);   according to Eq.(\ref{change logits PAFLDA})
                            \EndIf 
                        \EndIf
					\State Upload $Z_{k}$ to the server;
			\EndFor						
		\Until{Training stop}  ;
	\end{algorithmic}
\end{algorithm}

\begin{algorithm}[t]
	\renewcommand{\algorithmicrequire}{\textbf{Input:}}
	\renewcommand{\algorithmicensure}{\textbf{Output:}}
	\caption{FDLA or PCFDLA on the Server}
	\label{aggregation on server}
	\begin{algorithmic}[1] 
		\Repeat
		\Repeat
		\State Receive $Z_{k}$				
		\Until{Receive all $Z_{k}$ from K clients};  
		\State // aggregation process
		\State ZS $\leftarrow$ A($Z_{k}$) according to Eq(\ref{aggreation_logits});
		\State distribute SZ to all clients;
		\Until{Training stop};
	\end{algorithmic}
\end{algorithm}
\FloatBarrier

In summary, compared to FDLA, which involves the attacker making themselves "dumb" first to send misleading messages to the aggregation server, PCFDLA interferes with the FD process by cleverly manipulating logits values during the local client knowledge upload phase. It ensures that the most misleading answer for each new sample is sent to the aggregation server every time. This attack not only affects the learning of all clients, but more importantly, because PCFDLA retains its own recognition ability, it can make the incorrect answers even more misleading.

\subsection{Victims Oriented Metric in FD}
To evaluate the effectiveness of PCFDLA on FD systems, a rigorous assessment of its performance using standard and well-known metrics is required for fair comparison. The evaluation should cover multiple aspects to determine the system's effectiveness and practicality, including the average recognition accuracy of the models. Given the way FDLA attacks FD systems, certain metrics may be more important than others, such as the change in accuracy of the victims as a whole before and after the attack. To ensure scientific rigor, it is crucial to provide detailed metrics for effectively evaluating the proposed method.

FDLA has proposed several evaluation metrics to measure the accuracy and robustness of various misleading attacks. These metrics typically focused on the average accuracy of the entire model. We report the current metrics used to evaluate the performance of misleading attack methods on the FD system and introduce a new evaluation metric based on this: the change in accuracy of the victim group.

Specifically, the previous research work metrics mainly calculate the average model accuracy across various scenarios: different proportions of malicious attackers on various datasets, different data distributions, different numbers of clients, and different model structures.

The aim is minimize the the following objective function:
\begin{equation}
\label{old metric}
    TolAvgAcc= \frac{1}{K} \sum_{k=1}^{K} Acc
\end{equation}
where $TolAvgAcc$ represents the average accuracy of all models, $K$ denotes the total number of client devices, and $Acc$ represents the accuracy of each model.

However, these metrics do not directly show the impact on the victim group. In PCFDLA, by analyzing and comparing the accuracy changes of all victim models, we can more directly observe the interference of the attack on the performance of the victim models. Additionally, since PCFDLA has controllable attack intensity, we have added an evaluation of the impact of different attack intensities on both the original and new metrics in this paper. Specifically, we introduce the following five new metrics: the average model accuracy of all models and the average model accuracy of all victim models under different proportions of malicious attackers on various datasets, different data distributions, different numbers of clients, different model structures, and different attack intensities.

The aim is minimize the the following objective function:
\begin{equation}
\label{new metric}
    VctmAvgAcc= \frac{1}{KV} \sum_{kv=1}^{KV} Acc
\end{equation}
where $VctmAvgAcc$ represents the average accuracy of models from all victims, $KV$ denotes the total number of attacked client devices, and $Acc$ represents the accuracy of each model.

By introducing the above new metrics, we can more clearly show the impact of different attack methods on the target models.
\section{Experiments}
\subsection{Experimental Setup}
We conducted experiments on three common datasets: CINIC-10 \cite{cinic10}, CIFAR-10 \cite{krizhevsky2009learning}, and SVHN \cite{netzer2011reading}. Each dataset was divided into 50 non-i.i.d. (non-independent and identically distributed) parts, using the hyperparameter \( \alpha \) to control the degree of data heterogeneity, set to 1.0 \cite{wu2023fedcache}. We ensured that all models reached convergence. We evaluated our method under two FD frameworks, namely FedCache \cite{wu2023fedcache} and FD \cite{jeong2018communication}. We adopted the $A^C_1$ client model architecture from \cite{wu2023fedcache}. Other hyperparameters were consistent with those used in \cite{wu2023fedcache}.

\subsection{Baselines}
We compared our proposed PCFDLA algorithm with three baseline algorithms from FDLA that modify logits during training: Random poisoning randomly modifies logits, turning their values into random numbers between 0 and 1; Zero poisoning sets all logits to zero before uploading; Misleading attack (FDLA) uses a transformation mapping \(t\) that maps the element with the highest confidence to the position of the lowest confidence, while the remaining elements move up one position in the ranking.

\subsection{Results}
Based on the main experimental setup, the final results are shown in Tables \ref{main_experiment_1}, \ref{main_experiment_2}, and \ref{main_experiment_3}. The data indicate that under both FD and FedCache algorithms, whether on the SVHN, CIFAR-10, or CINIC-10 datasets, the PCFDLA attack has the most significant impact on model accuracy in all scenarios, causing an accuracy loss of up to 5\% - 20\%. In contrast, the attack strategies proposed in this paper (random poisoning, zero poisoning, and misleading attacks) do not significantly affect the model. In some cases, they can even slightly improve the model’s recognition rate. As the proportion of malicious attackers increases, the effectiveness of PCFDLA further strengthens, reducing the model recognition accuracy to around 20\%. In contrast, the other three baseline algorithms show no noticeable change in model accuracy interference. In the FD algorithm, with 30\% malicious attackers, PCFDLA proves to be the most effective, achieving the lowest recognition accuracy of 31.22\%. Similarly, in the FedCache algorithm, with 30\% malicious attackers, PCFDLA also proves to be the most effective, achieving the lowest recognition accuracy of 26.57\%.
\vspace{40pt}
\begin{table}[!htbp]
\centering
\caption{The average recognition accuracy (\%) of all models on the SVHN dataset before and after different attacks under FD and FedCache algorithms, as well as the accuracy changes of the victim models before and after the attacks, with the \uline{lowest} and \textbf{second lowest} accuracy values marked.}
\label{main_experiment_1}
\begin{tabular}{c|>{\centering\arraybackslash}p{1.5cm}|>{\centering\arraybackslash}p{1.5cm}|>{\centering\arraybackslash}p{1.5cm}|>{\centering\arraybackslash}p{1.5cm}|>{\centering\arraybackslash}p{1.5cm}|>{\centering\arraybackslash}p{1.5cm}}
\hline
\multirow{3}{*}{Poisoning Methods} & \multicolumn{6}{c}{FD} \\ \cline{2-7} 
 & \multicolumn{3}{c|}{Average Accuracy} & \multicolumn{3}{c}{Victims Average accuracy } \\ \cline{2-7} 
 & 10\% & 20\% & 30\% & 10\% & 20\% & 30\% \\ \hline
No Poisoning & \multicolumn{3}{c|}{66.51} & \multicolumn{3}{c}{66.51} \\ \hline
Random Poisoning & 66.07 & 65.36 & 66.84 & 66.58 & 67.21 & 68.90 \\ \hline
Zero Poisoning & 65.78 & 65.88 & {\ul 65.57} & 66.19 & {\ul 66.94} & 67.97 \\ \hline
FDLA & {\ul 65.23} & {\ul 66.39} & 65.84 & {\ul 65.74} & 67.70 & 68.46 \\ \hline
\textbf{PCFDLA} & \textbf{59.02} & \textbf{33.37} & \textbf{40.11} & \textbf{58.75} & \textbf{33.90} & \textbf{43.78} \\ \hline
\multirow{3}{*}{Poisoning Methods} & \multicolumn{6}{c}{FedCache} \\ \cline{2-7} 
 & \multicolumn{3}{c|}{Average Accuracy} & \multicolumn{3}{c}{Victims Average accuracy } \\ \cline{2-7} 
 & 10\% & 20\% & 30\% & 10\% & 20\% & 30\% \\ \hline
No Poisoning & \multicolumn{3}{c|}{44.27} & \multicolumn{3}{c}{44.27} \\ \hline
Random Poisoning & {\ul 44.10} & 43.99 & 43.75 & {\ul 44.80} & 45.83 & 45.95 \\ \hline
Zero Poisoning & 44.52 & 44.08 & 43.39 & 45.20 & 46.14 & 45.86 \\ \hline
FDLA & 44.32 & {\ul 42.76} & {\ul 41.21} & 45.04 & {\ul 44.81} & {\ul 43.74} \\ \hline
\textbf{PCFDLA} & \textbf{32.28} & \textbf{27.51} & \textbf{26.57} & \textbf{33.62} & \textbf{28.28} & \textbf{26.94} \\ \hline
\end{tabular}
\end{table}

\begin{table}[!htbp]
\centering
\caption{The average recognition accuracy (\%) of all models on the CIFAR-10 dataset before and after different attacks under FD and FedCache algorithms, as well as the accuracy changes of the victim models before and after the attacks, with the \uline{lowest} and \textbf{second lowest} accuracy values marked.}
\label{main_experiment_2}
\begin{tabular}{c|>{\centering\arraybackslash}p{1.5cm}|>{\centering\arraybackslash}p{1.5cm}|>{\centering\arraybackslash}p{1.5cm}|>{\centering\arraybackslash}p{1.5cm}|>{\centering\arraybackslash}p{1.5cm}|>{\centering\arraybackslash}p{1.5cm}}
\hline
\multirow{3}{*}{Poisoning Methods} & \multicolumn{6}{c}{FD} \\ \cline{2-7} 
 & \multicolumn{3}{c|}{Average Accuracy} & \multicolumn{3}{c}{Victims Average accuracy } \\ \cline{2-7} 
 & 10\% & 20\% & 30\% & 10\% & 20\% & 30\% \\ \hline
No Poisoning & \multicolumn{3}{c|}{56.08} & \multicolumn{3}{c}{56.08} \\ \hline
Random Poisoning & 56.54 & 56.38 & {\ul 56.13} & 57.18 & {\ul 57.71} & {\ul 57.04} \\ \hline
Zero Poisoning & 56.91 & {\ul 56.20} & 56.53 & 57.29 & 57.76 & 57.67 \\ \hline
FDLA & {\ul 56.31} & 56.68 & 56.96 & {\ul 57.09} & 58.33 & 57.82 \\ \hline
\textbf{PCFDLA} & \textbf{52.59} & \textbf{43.97} & \textbf{39.69} & \textbf{53.05} & \textbf{45.15} & \textbf{41.15} \\ \hline
\multirow{3}{*}{Poisoning Methods} & \multicolumn{6}{c}{FedCache} \\ \cline{2-7} 
 & \multicolumn{3}{c|}{Average Accuracy} & \multicolumn{3}{c}{Victims Average accuracy } \\ \cline{2-7} 
 & 10\% & 20\% & 30\% & 10\% & 20\% & 30\% \\ \hline
No Poisoning & \multicolumn{3}{c|}{52.98} & \multicolumn{3}{c}{52.98} \\ \hline
Random Poisoning & 52.24 & 52.32 & 52.05 & 52.82 & 53.62 & 53.02 \\ \hline
Zero Poisoning & 51.92 & 52.42 & 52.04 & 52.48 & 53.72 & 53.37 \\ \hline
FDLA & {\ul 51.56} & {\ul 51.91} & {\ul 50.22} & {\ul 52.01} & {\ul 53.48} & {\ul 51.52} \\ \hline
\textbf{PCFDLA} & \textbf{39.27} & \textbf{36.73} & \textbf{26.89} & \textbf{39.76} & \textbf{37.69} & \textbf{24.91} \\ \hline
\end{tabular}
\end{table}

\begin{table}[!ht]
\centering
\caption{The average recognition accuracy (\%) of all models on the CINIC-10 dataset before and after different attacks under FD and FedCache algorithms, as well as the accuracy changes of the victim models before and after the attacks, with \uline{lowest} and \textbf{second lowest} accuracy values marked.}
\label{main_experiment_3}
\begin{tabular}{c|>{\centering\arraybackslash}p{1.5cm}|>{\centering\arraybackslash}p{1.5cm}|>{\centering\arraybackslash}p{1.5cm}|>{\centering\arraybackslash}p{1.5cm}|>{\centering\arraybackslash}p{1.5cm}|>{\centering\arraybackslash}p{1.5cm}}
\hline
\multirow{3}{*}{Poisoning Methods} & \multicolumn{6}{c}{FD} \\ \cline{2-7} 
 & \multicolumn{3}{c|}{Average Accuracy} & \multicolumn{3}{c}{Victims Average accuracy } \\ \cline{2-7} 
 & 10\% & 20\% & 30\% & 10\% & 20\% & 30\% \\ \hline
No Poisoning & \multicolumn{3}{c|}{53.15} & \multicolumn{3}{c}{53.15} \\ \hline
Random Poisoning & 53.36 & 52.94 & 52.77 & 53.89 & 54.51 & {\ul 54.11} \\ \hline
Zero Poisoning & {\ul 52.61} & {\ul 52.57} & 52.90 & {\ul 53.15} & {\ul 54.07} & 54.30 \\ \hline
FDLA & 53.00 & 52.96 & {\ul 52.70} & 53.35 & 54.41 & 54.27 \\ \hline
\textbf{PCFDLA} & \textbf{48.03} & \textbf{44.13} & \textbf{31.22} & \textbf{48.28} & \textbf{45.48} & \textbf{32.07} \\ \hline
\multirow{3}{*}{Poisoning Methods} & \multicolumn{6}{c}{FedCache} \\ \cline{2-7} 
 & \multicolumn{3}{c|}{Average Accuracy} & \multicolumn{3}{c}{Victims Average accuracy } \\ \cline{2-7} 
 & 10\% & 20\% & 30\% & 10\% & 20\% & 30\% \\ \hline
No Poisoning & \multicolumn{3}{c|}{46.67} & \multicolumn{3}{c}{49.40} \\ \hline
Random Poisoning & 46.81 & 46.84 & 46.56 & {\ul 47.19} & 48.31 & 48.30 \\ \hline
Zero Poisoning & {\ul 47.43} & 46.71 & 46.62 & 47.91 & 48.17 & {\ul 48.20} \\ \hline
FDLA & 47.66 & {\ul 46.55} & {\ul 46.53} & 48.05 & {\ul 48.10} & 48.60 \\ \hline
\textbf{PCFDLA} & \textbf{35.70} & \textbf{33.51} & \textbf{27.76} & \textbf{36.13} & \textbf{34.81} & \textbf{27.96} \\ \hline
\end{tabular}
\end{table}

\vspace{90pt}
\begin{figure}[!htbp]
	\subfigure
	{
		\begin{minipage}[b]{0.5\linewidth}
			\centering
			\includegraphics[width=\textwidth,height= 4.76cm]{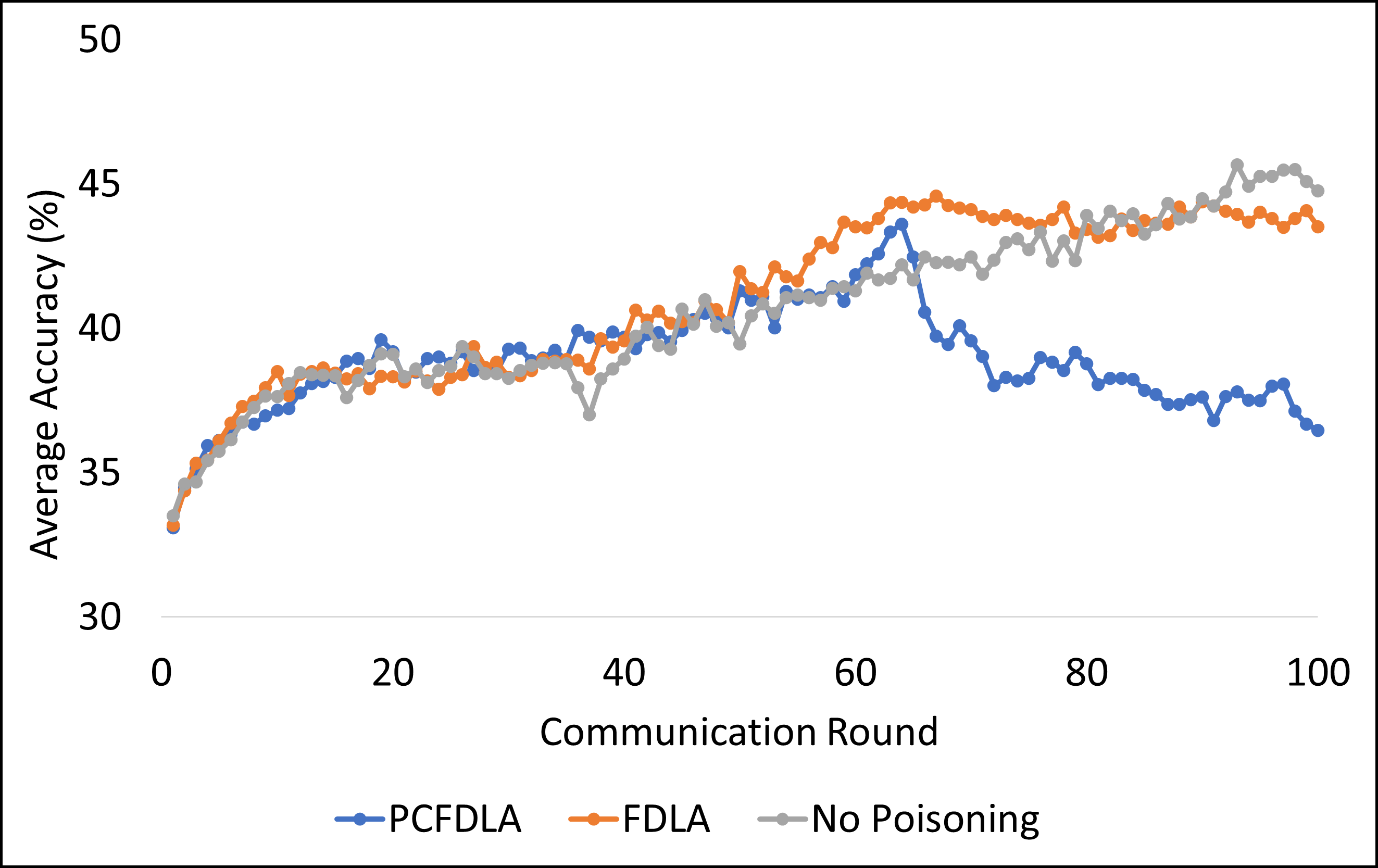}
		\end{minipage}
	}
	\subfigure
	{
		\begin{minipage}[b]{0.5\linewidth}
			\centering
			\includegraphics[width=\textwidth,height= 4.76cm]{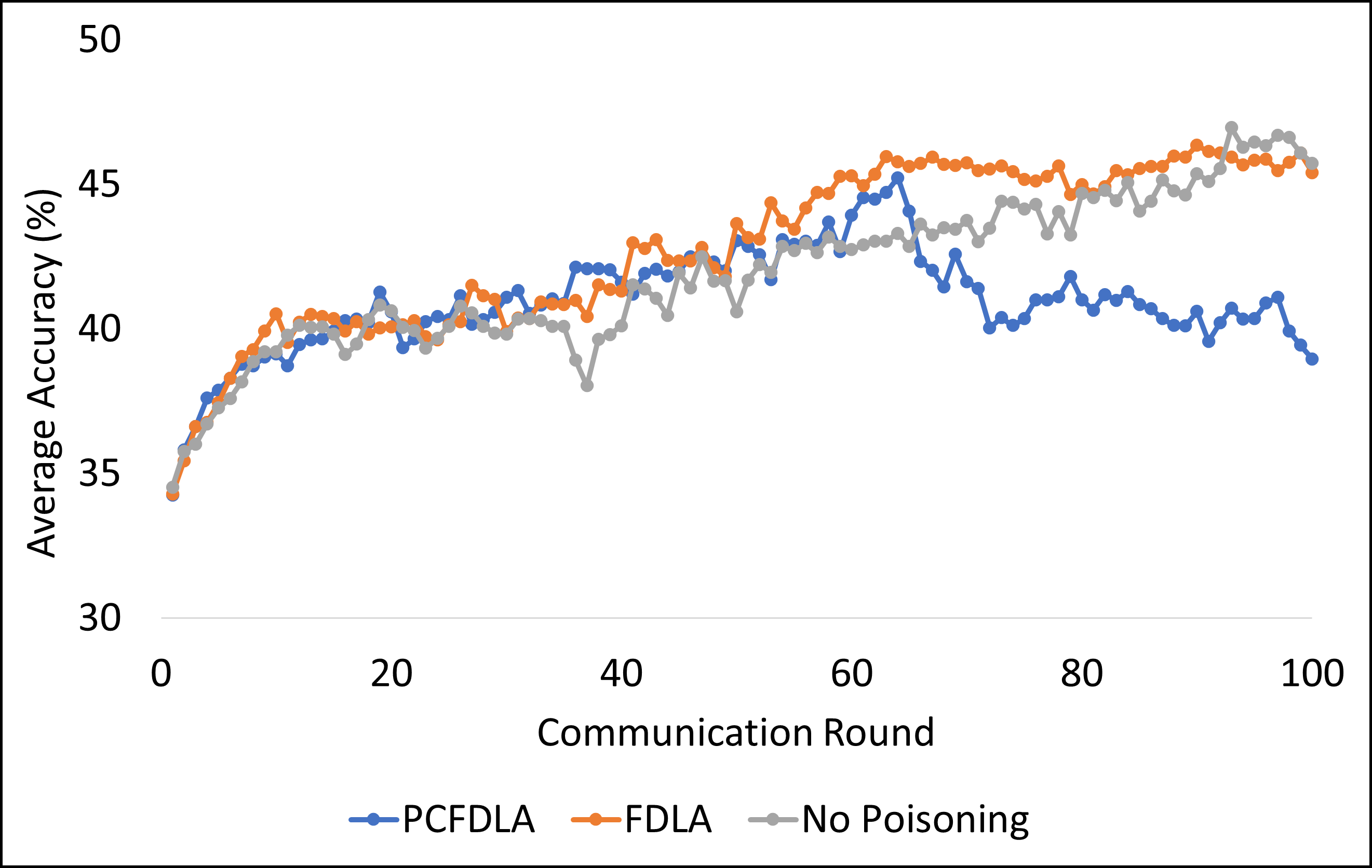}
		\end{minipage}
	}	
        \caption{Convergence impact of three attack types on FD and FedCache in CIFAR-10 with 30\% attackers, tracking average accuracy per communication round.}
        \label{impact on convenience}
 \end{figure}
\FloatBarrier
\vspace{90pt}
\begin{figure}[!htbp]
    \centering
    \subfigure[10\% malicious clients]
    {
        \begin{minipage}{5cm}
            \centering
            \includegraphics[width=5cm]{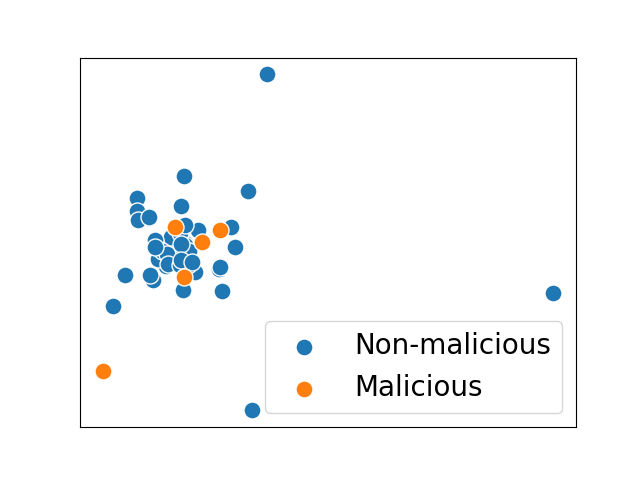}
        \end{minipage}
    }
    \subfigure[20\% malicious clients]
    {
        \begin{minipage}{5cm}
            \centering
            \includegraphics[width=5cm]{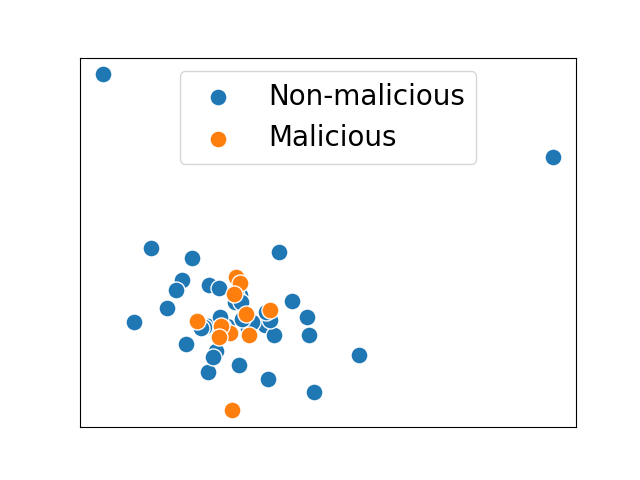}
        \end{minipage}
    }
    \subfigure[30\% malicious clients]
    {
        \begin{minipage}{5cm}
            \centering
            \includegraphics[width=5cm]{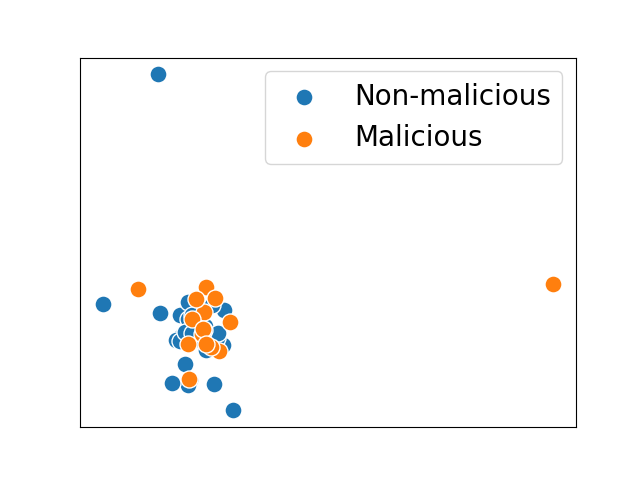}
        \end{minipage}
    }
    \caption{Illustration of the distribution of malicious and non-malicious user models after PCA dimensionality reduction. The malicious users are marked in blue, while the non-malicious users are marked in orange.}
    \label{The Stealthiness nature of FDLA}
\end{figure}
\begin{figure}[!htbp]
	\subfigure
	{
		\begin{minipage}[b]{0.5\linewidth}
			\centering
			\includegraphics[width=\textwidth,height=0.2\textheight]{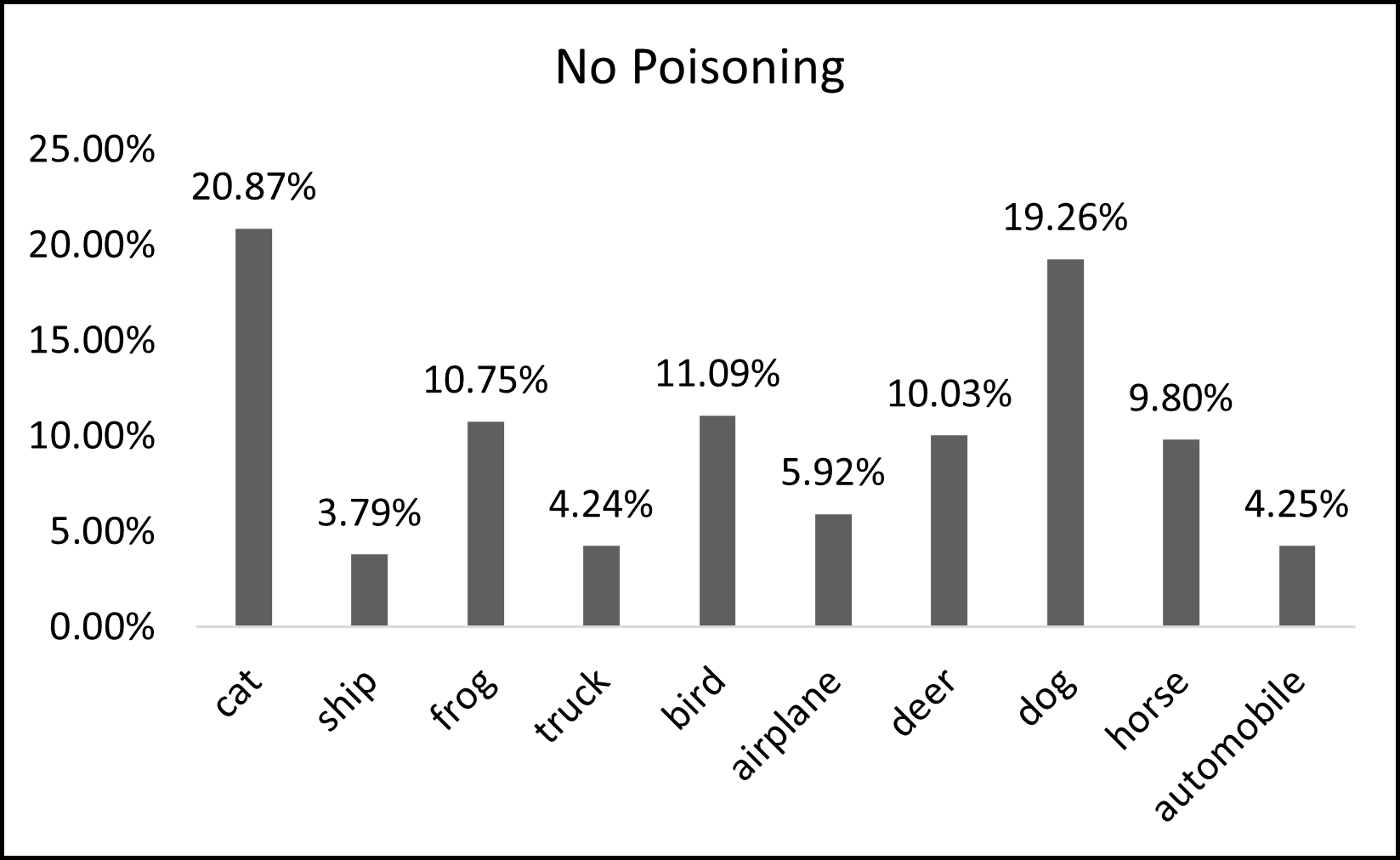}
			\label{no impact on catset in FedCache}
		\end{minipage}

	}
	\subfigure
	{
		\begin{minipage}[b]{0.5\linewidth}
			\centering
			\includegraphics[width=\textwidth,height=0.2\textheight]{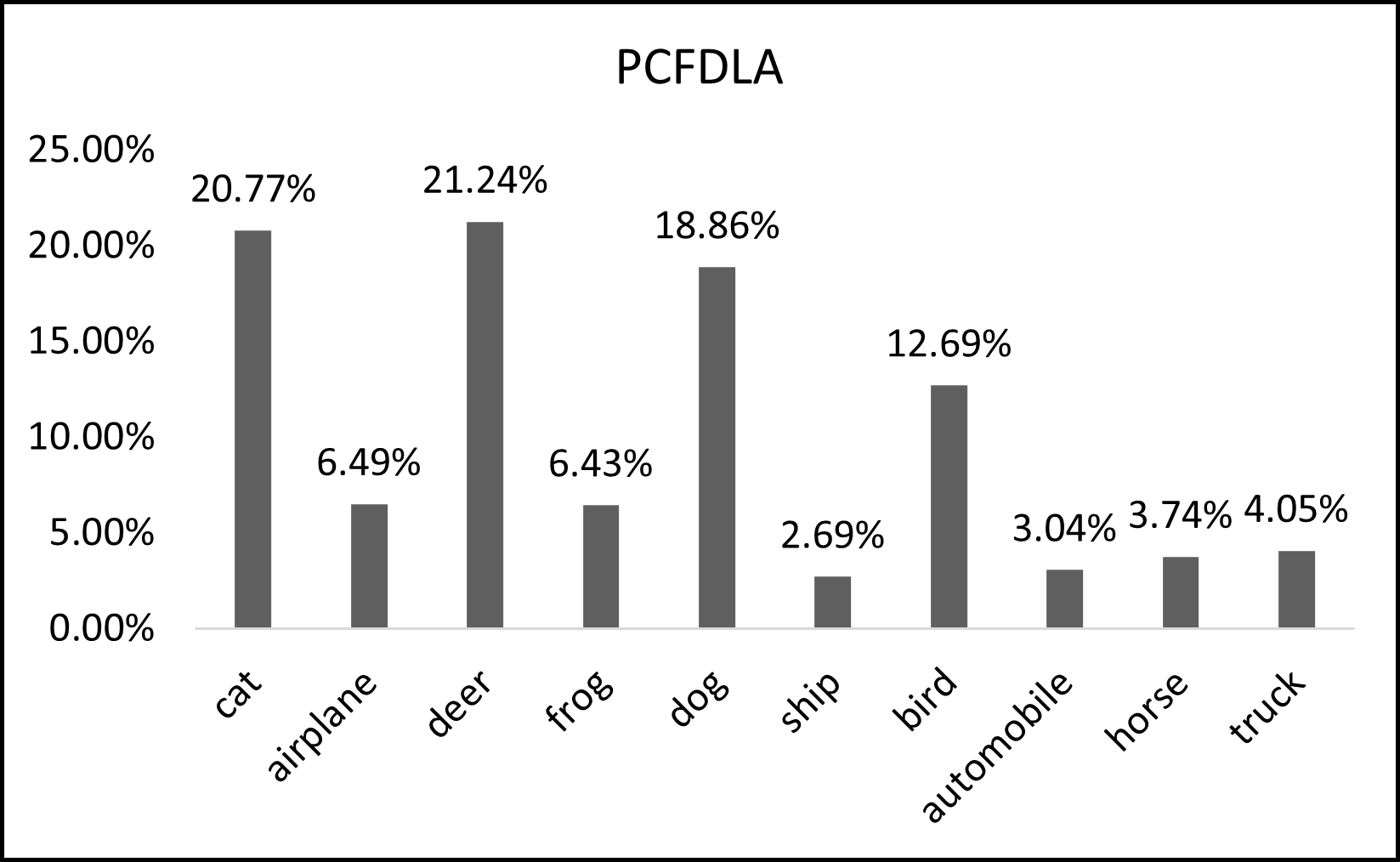}
			\label{impact on catset in FedCache}
		\end{minipage}
	}	
	\caption{Illustration of the misleading effect of PCFDLA to client models. The vertical axis of the chart represents the probability statistics of the test results for cat samples, while the horizontal axis represents the categories of the test results.}
	\label{impact on catset}
 \end{figure}

\FloatBarrier

Referring to Figure \ref{impact on convenience}, the results indicate that although FDLA generally reduces the average recognition accuracy of the model, it does not significantly impact the victims. In contrast, with PCFDLA, the experimental results show an accuracy loss of nearly 10\% for both the overall model and specifically for all victims.

Furthermore, by testing the cat test set in the CIFAR-10 dataset, the results in Figure \ref{impact on catset} (the left figure shows the predictive statistics of all models on the test set without tampering, while the right figure displays the statistical results after PCFDLA manipulation) show that after 40 rounds of misleading attacks, the number of times the model misclassified 50,000 cat samples as deer increased from 5,014 to 10,620, growing from 10\% to 20\%.

Finally, using PCA to validate the stealthiness of the attack in one of the experiments under the FedCache algorithm, the results in Figure \ref{The Stealthiness nature of FDLA} show that as the number of malicious attackers increases, the difference between their models and those of the victims is not significant. This proves that PCFDLA has good stealthiness.

Overall, considering all the average accuracy and the accuracy changes of non-malicious participants, PCFDLA is the most effective method overall.

\subsection{Ablation Study}
The ablation study is divided into three parts, aiming to explore the relationship between hyperparameters and performance: The experimental default configuration is the same as the main experiment, but only the SVHN dataset is used. The basic FD algorithm used is FedCache, and the poisoning ratio is set to 20 \%.
\begin{itemize}
	\item
	The first part examines the robustness of PCFDLA concerning data distribution:\\
    This experiment compares the effects of three different poisoning methods by setting $\alpha$ to three different values: 0.5, 1.0, and 3. This allows us to explore how the effectiveness of PCFDLA is influenced by the distribution of data among clients.

	\item
	The second part investigates the robustness of PCFDLA concerning the number of participating clients:\\
    This experiment sets up scenarios with different numbers of participants, specifically 10, 50, and 200, to compare the effects of three different poisoning methods. This investigation aims to understand how the effectiveness of FDLA is influenced by the number of clients in the federated learning network.

	\item
	The third part of the study explores the robustness of PCFDLA concerning model architecture: \\
    This experiment evaluates the performance of three different poisoning methods when configured with homogeneous or heterogeneous models to further study the robustness of PCFDLA to model structure. The experiment considers the homogeneity and heterogeneity of client models. Specifically, in the homogeneous model experiment, all clients use the same model architecture, denoted as \( A_{1}^{C} \). In the heterogeneous model case, FedCache assigns different model architectures to clients based on the client index modulo 3. When the index modulo 3 equals 0, 1, and 2, the assigned model architectures are \( A_{1}^{C} \), \( A_{2}^{C} \), and \( A_{3}^{C} \), respectively. That is:
	\begin{equation}\label{distribute_model}
		A_i^{C} = A_{(i \bmod 3) + 1}^{C}
	\end{equation}
\end{itemize}

\begin{table}[!htbp]
\centering
\caption{The impact of poisoning on model accuracy(\%) using the FedCache algorithm with a 20\% attack ratio and different data heterogeneity levels on the SVHN dataset, indicating \textbf{lowest} and \uline{second lowest} accuracy values.}
\label{different_partition_alpha}
\begin{tabular}{c|>{\centering\arraybackslash}p{1.5cm}|>{\centering\arraybackslash}p{1.5cm}|>{\centering\arraybackslash}p{1.5cm}|>{\centering\arraybackslash}p{1.5cm}|>{\centering\arraybackslash}p{1.5cm}|>{\centering\arraybackslash}p{1.5cm}}
\hline
\multirow{2}{*}{Poisoning Methods} & \multicolumn{3}{c|}{Average Accuracy} & \multicolumn{3}{c}{Victims Average accuracy } \\ \cline{2-7} 
 & \multicolumn{1}{c|}{$\alpha$=0.5} & \multicolumn{1}{c|}{$\alpha$=1.0} & $\alpha$=3.0 & \multicolumn{1}{c|}{$\alpha$=0.5} & \multicolumn{1}{c|}{$\alpha$=1.0} & $\alpha$=3.0 \\ \hline
No Poisoning & \multicolumn{1}{c|}{55.85} & \multicolumn{1}{c|}{44.27} & 37.36 & \multicolumn{1}{c|}{56.59} & \multicolumn{1}{c|}{44.27} & 37.28 \\ \hline
Random Poisoning & \multicolumn{1}{c|}{55.22} & \multicolumn{1}{c|}{43.99} & {\ul 35.57} & \multicolumn{1}{c|}{56.42} & \multicolumn{1}{c|}{45.83} & {\ul 35.15} \\ \hline
Zero Poisoning & \multicolumn{1}{c|}{55.32} & \multicolumn{1}{c|}{44.08} & 35.97 & \multicolumn{1}{c|}{56.55} & \multicolumn{1}{c|}{46.14} & 35.68 \\ \hline
FDLA & \multicolumn{1}{c|}{{\ul 52.77}} & \multicolumn{1}{c|}{{\ul 42.76}} & 35.73 & \multicolumn{1}{c|}{{\ul 54.39}} & \multicolumn{1}{c|}{{\ul 44.81}} & 36.10 \\ \hline
\textbf{PCFDLA} & \multicolumn{1}{c|}{\textbf{37.72}} & \multicolumn{1}{c|}{\textbf{27.51}} & \textbf{22.62} & \multicolumn{1}{c|}{\textbf{38.51}} & \multicolumn{1}{c|}{\textbf{28.28}} & \textbf{22.59} \\ \hline
\end{tabular}
\end{table}
\vspace{13pt}

\begin{table}[!htbp]
\centering
\caption{The impact of the number of clients on the accuracy (\%) with a 20\% poisoning ratio using FedCache on the SVHN dataset, indicating \textbf{lowest} and \uline{second lowest} accuracy values.}
\label{different_client_number}
\begin{tabular}{c|>{\centering\arraybackslash}p{1.5cm}|>{\centering\arraybackslash}p{1.5cm}|>{\centering\arraybackslash}p{1.5cm}|>{\centering\arraybackslash}p{1.5cm}|>{\centering\arraybackslash}p{1.5cm}|>{\centering\arraybackslash}p{1.5cm}}
\hline
\multirow{2}{*}{Poisoning Methods} & \multicolumn{3}{c|}{Average Accuracy} & \multicolumn{3}{c}{Victims Average accuracy } \\ \cline{2-7} 
 & \multicolumn{1}{c|}{10 clients} & \multicolumn{1}{c|}{50 clients} & 200 clients & \multicolumn{1}{c|}{10 clients} & \multicolumn{1}{c|}{50 clients} & 200 clients \\ \hline
No Poisoning & \multicolumn{1}{c|}{69.03} & \multicolumn{1}{c|}{44.27} & 33.24 & \multicolumn{1}{c|}{65.55} & \multicolumn{1}{c|}{44.27} & 33.06 \\ \hline
Random Poisoning & \multicolumn{1}{c|}{70.62} & \multicolumn{1}{c|}{43.99} & 32.51 & \multicolumn{1}{c|}{68.64} & \multicolumn{1}{c|}{45.83} & 32.70 \\ \hline
Zero Poisoning & \multicolumn{1}{c|}{{\ul 69.63}} & \multicolumn{1}{c|}{44.08} & 31.84 & \multicolumn{1}{c|}{{\ul 68.02}} & \multicolumn{1}{c|}{46.14} & 31.97 \\ \hline
FDLA & \multicolumn{1}{c|}{71.09} & \multicolumn{1}{c|}{{\ul 42.76}} & {\ul 30.26} & \multicolumn{1}{c|}{69.47} & \multicolumn{1}{c|}{{\ul 44.81}} & {\ul 30.70} \\ \hline
\textbf{PCFDLA} & \multicolumn{1}{c|}{\textbf{55.43}} & \multicolumn{1}{c|}{\textbf{27.51}} & \textbf{19.95} & \multicolumn{1}{c|}{\textbf{57.19}} & \multicolumn{1}{c|}{\textbf{28.28}} & \textbf{20.47} \\ \hline
\end{tabular}
\end{table}

\vspace{13pt}

\begin{table}[!htbp]
\centering
\caption{The impact of model structure on the accuracy (\%) with a 20\% poisoning ratio using FedCache on the SVHN dataset, highlighting the \textbf{lowest} and \uline{second lowest} accuracy values.}
\label{different_heterogeneity}
\begin{tabular}{c|>{\centering\arraybackslash}p{2.5cm}|>{\centering\arraybackslash}p{2.5cm}|>{\centering\arraybackslash}p{2.5cm}|>{\centering\arraybackslash}p{2.5cm}}
\hline
\multirow{2}{*}{Poisoning Methods} & \multicolumn{2}{c|}{Average Accuracy} & \multicolumn{2}{c}{Victims Average accuracy } \\ \cline{2-5} 
 & \multicolumn{1}{c|}{Homogeneous} & Heterogeneous & \multicolumn{1}{c|}{Homogeneous} & Heterogeneous \\ \hline
No Poisoning & \multicolumn{1}{c|}{44.27} & 53.79 & \multicolumn{1}{c|}{44.27} & 55.50 \\ \hline
Random Poisoning & \multicolumn{1}{c|}{43.99} & 52.29 & \multicolumn{1}{c|}{45.83} & 54.19 \\ \hline
Zero Poisoning & \multicolumn{1}{c|}{44.08} & 53.07 & \multicolumn{1}{c|}{46.14} & 54.90 \\ \hline
FDLA & \multicolumn{1}{c|}{{\ul 42.76}} & {\ul 51.19} & \multicolumn{1}{c|}{{\ul 44.81}} & {\ul 53.72} \\ \hline
\textbf{PCFDLA} & \multicolumn{1}{c|}{\textbf{27.51}} & \textbf{33.61} & \multicolumn{1}{c|}{\textbf{28.28}} & \textbf{35.55} \\ \hline
\end{tabular}
\end{table}

\begin{figure}[!htbp]
    \centering
    \includegraphics[scale = 0.475]{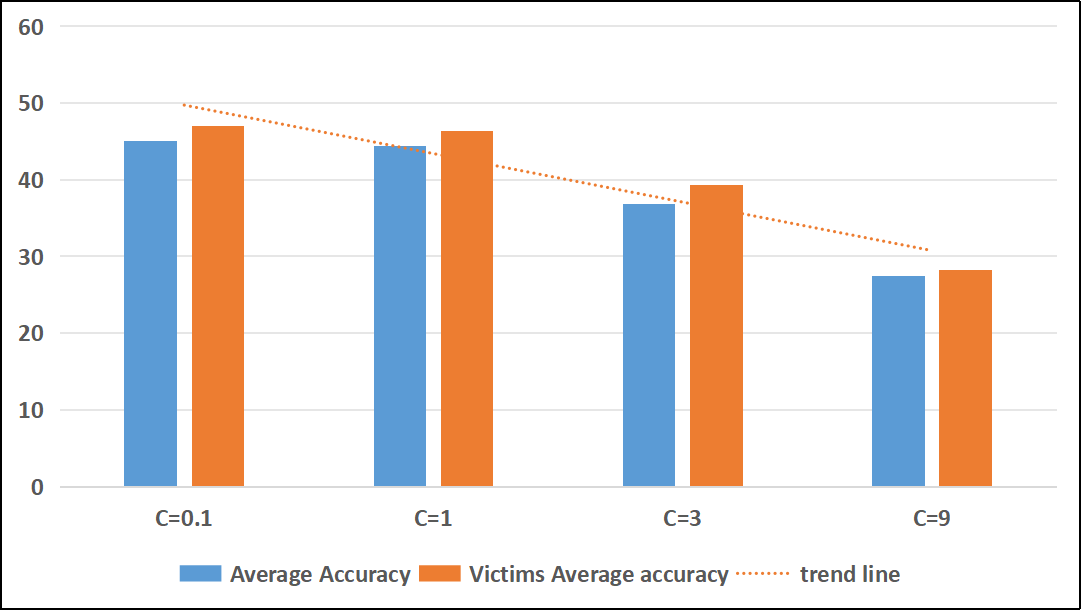}
    \caption{Exploring the impact of the constant C for logits control on performance.}
    \label{different_C}
\end{figure}

Table \ref{different_partition_alpha} shows the impact of data distribution on PCFDLA, FDLA, random poisoning, and zero poisoning. The results indicate that with increased heterogeneity, the impact of FDLA, random poisoning, and zero poisoning on model accuracy remains around 2\%, while PCFDLA further decreases the recognition rate from 37\% to a minimum of 22\%, proving to be the most effective method. As the diversity of data types in the dataset increases, FDLA, random poisoning, and zero poisoning maintain almost the same level of deception, indirectly proving the robustness of PCFDLA to data heterogeneity.

Table \ref{different_client_number} shows the impact of the number of participants on PCFDLA, FDLA, random poisoning, and zero poisoning. The results indicate that with an increase in the number of clients, the impact of FDLA, random poisoning, and zero poisoning on model accuracy slightly increases, with PCFDLA again proving to be the most effective, maintaining about a 10\% model accuracy loss.

Table \ref{different_heterogeneity} shows the impact of model structure on PCFDLA, FDLA, random poisoning, and zero poisoning. The results show that regardless of whether the models are heterogeneous or homogeneous, PCFDLA causes a model accuracy decline of about 20\%, indicating that PCFDLA has a certain robustness to model structure.

Figure \ref{different_C} shows the impact of the hyperparameter $C$ value of PCFDLA on model accuracy. The results indicate that as C increases, PCFDLA causes the model accuracy to gradually decline by about 20\% , showing that the hyperparameter $C$ of PCFDLA has a controlling effect on the strength of the attack.
\section{Conclusion}
This paper introduces a novel logits poisoning attack method, PCFDLA, for federated distillation. PCFDLA modifies the logits sent from the client to the server, guiding the model to generate high-confidence incorrect predictions. It can control the strength of the attack, thereby interfering with model training at different intensities. During Federated Distillation, PCFDLA subtly influences model training by redesign the knowledge-sharing mechanism. Experimental results demonstrate the strong attack effect of PCFDLA across various settings.


\bmsection*{Acknowledgments}
This work was supported in part by the Fundamental Research Funds for the Central Universities under Grant 2021JBM008 and Grant 2022JBXT001, and in part by the National Natural Science Foundation of China (NSFC) under Grant 61872028. (Corresponding author: Bo Gao.) 



\bmsection*{Conflict of interest}
The authors declare no potential conflict of interests.


\end{document}